\documentclass{vgtc}                          

\graphicspath{{figures/}{pictures/}{images/}{./}} 

\usepackage{times}                     

\usepackage{tabu}                      
\usepackage{booktabs}                  
\usepackage{lipsum}                    
\usepackage{mwe}                       

\usepackage{mathptmx}                  

\onlineid{0}

\vgtccategory{Research}

\vgtcinsertpkg

\title{Visual Fingerprints for LLM Generation Comparison}

\author{Amal Alnouri\thanks{e-mail: amal.alnouri@jku.at}\\ %
        \scriptsize Johannes Kepler University Linz %
\and Andreas Hintereiter\thanks{e-mail: andreas.hinterreiter@jku.at}\\ %
        \scriptsize Johannes Kepler University Linz %
\and Christina Humer\thanks{e-mail: christina.humer@inf.ethz.ch}\\ %
        \scriptsize ETH Zürich %
\and Furui Cheng\\ %
        \scriptsize Independent Researcher %
\and Marc Streit\thanks{e-mail: marc.streit@jku.at}\\ %
        \scriptsize Johannes Kepler University Linz }

\teaser{%
  \centering
  \includegraphics[width=0.8\linewidth, alt={The computational pipeline and interface of the proposed system.}]{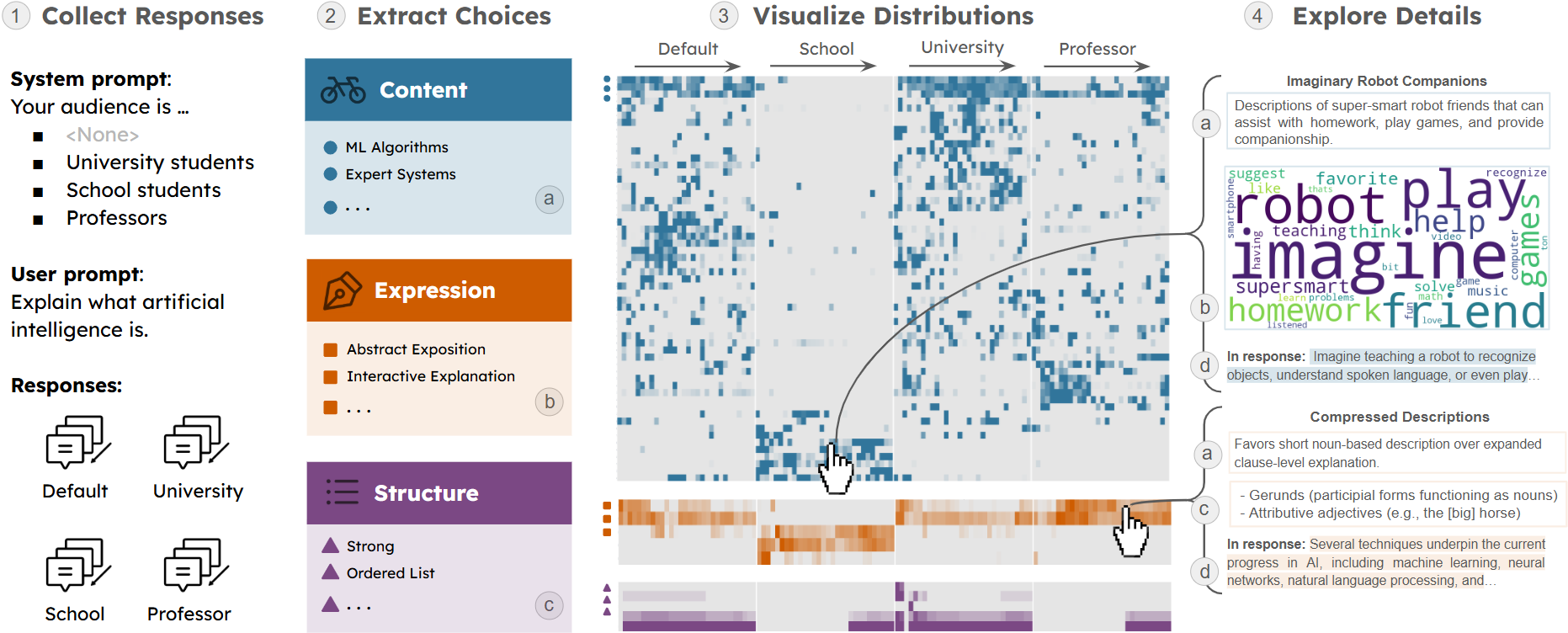}
  \caption{%
    The fingerprinting pipeline collects LLM responses under different generation conditions (e.g., varying system prompts), defines a shared choice space across three dimensions---\emph{content}, \emph{expression}, and \emph{structure}---and visualizes their distributions as comparable fingerprints.
  }
  \label{fig:teaser}
}

\abstract{%
Large language model (LLM) outputs arise from complex interactions among prompts, system instructions, model parameters, and architecture. 
We refer to specific configurations of these factors as generation conditions, each of which can bias outputs in various ways.
Understanding how different generation conditions shape model behaviors is essential for tasks such as prompt design and model evaluation, yet it remains challenging due to the stochastic and open-ended nature of text generation.
We present an approach to visually compare LLM outputs across generation conditions by modeling responses as collections of linguistic choices, including content, expression, and structure. We extract these choices using natural language processing pipelines and represent their distributions across repeated samples. We then visualize these distributions as visual fingerprints, enabling direct, distribution-level comparison of condition-specific tendencies.
Through four usage scenarios, we demonstrate how visual fingerprints reveal consistent patterns in LLM behavior that are difficult to observe through individual responses or aggregate metrics.
} 


\graphicspath{{figs/}{figures/}{pictures/}{images/}{./}} 

\usepackage{tabu}                      
\usepackage{booktabs}                  
\usepackage{lipsum}                    
\usepackage{mwe}                       
\usepackage{ccicons}                   

\usepackage{mathptmx}                  

\usepackage{csquotes}
\usepackage{cleveref}

\usepackage[loadonly]{enumitem}

\newcommand{\citet}[2]{#2~et~al.~\cite{#1}}

\newcommand{\llmquote}[1]{%
    \textit{\enquote{#1}}%
}

\makeatletter

\newlength{\dreqlabellength}
\newlist{dreq@ls}{enumerate}{1}
\setlist[dreq@ls]{%
    itemsep=.2ex,
    topsep=.2ex,
    align=left,
    labelindent=0pt,
    labelwidth=\dreqlabellength,
    labelsep*=.2em,
    leftmargin =!,
    label=\textbf{DR\arabic*}
}
    {\end{dreq@ls}}

\makeatother

\newcommand{\transition}[2]{%
  \vspace{0.5em}%
  \noindent\textbf{#1.} #2%
}

\begin{document}


\firstsection{Introduction}

\maketitle

Large Language Models (LLMs) have become powerful tools for generating text in response to human-written natural-language instructions. This output reflects complex interactions between prompts and model settings.
Understanding how LLM outputs vary is critical in scenarios such as model evaluations and prompt engineering~\cite{helm2023, NEURIPS2024_ad236edc}. 
Users and practitioners frequently experiment with different prompts and system settings to evaluate and understand model behaviors.
However, they lack tools to systematically compare the resulting outputs.
Without such support, selecting between alternatives becomes a trial-and-error process.

Comparing open-ended output is challenging. 
First, stochasticity in model generations implies that a single response is not sufficient to understand the broad distribution of the possible outputs~\cite{parrot2021, wadi-fredette-2025-monte, zhang2025beyond}. 
Second, in open-ended generation tasks, evaluation is subjective, lacking clear ground-truth criteria~\cite{illdefined2026}. 
Third, textual data is hard to compare at scale. Making sense of a long piece of text requires sequential reading and manual inspections, making it tedious and hard to identify patterns across multiple responses. 

In this work, we propose an approach for visually comparing LLM outputs under different generation conditions. 
Despite the variability across individual outputs, LLMs tend to make recurring choices across repeated generations \cite{plagiarism2023, chang-bergen-2024-language}. Therefore, in our approach, we transform repeated LLM generations into comparable visual fingerprints, where each fingerprint summarizes the distribution of linguistic choices made under one generation condition.
We employ natural language processing (NLP) pipelines to extract linguistic choices from LLM responses, including addressed topics, communication styles, and structural formats. 
We then visualize their distributions as heatmaps, enabling direct comparison of condition-specific tendencies.
We discuss insights that emerge from applying our approach in four usage scenarios. We compare (\textit{i})~different LLM models for advice-giving, (\textit{ii})~prompt variations with varying politeness levels, (\textit{iii})~different LLM personas, and (\textit{iv})~the style of two LLM models with that of humans in storytelling.

In summary, our contributions are: (\textit{1})~a choice-based representation of LLM response distributions that models outputs through content, expression, and structure choices; (\textit{2})~a visual fingerprint design that enables condition-level comparison of stochastic open-ended generations; (\textit{3})~four usage scenarios demonstrating how the approach reveals patterns induced by different conditions.

\section{Related Work}

\transition{Visual Text Comparison}Prior literature discusses various visual approaches for document-level comparison. For text reuse, fine-grained text units such as words, n-grams, and sentences have been directly compared using sequence-aligned heatmaps \cite{yousef2020survey, ribler2000plagiarism}.  One axis lists documents, with individual text units forming the cell, and their sequential order mapped along the other axis. Grid-based heatmaps \cite{yousef2020survey, janicke2014reuse} have also been employed in this context to encode aggregated alignments between document pairs, where the entire document set is organized along both axes. For semantic and stylistic similarity analysis, previous work overlays each document with a text-oriented heatmap \cite{yousef2020survey, Weber2007colors, keim2007fingerprint} indicating the presence or intensity of a specific linguistic feature. Consistent word clouds have also been proposed to detect document-level similarity \cite{castella2014storm}. Comparing collections of documents has also been explored. For instance, Scattertext \cite{kessler-2017-scattertext} proposes divergent scatter plots where each point represents a term, supporting contrastive analysis between two document collections. Likewise, TextDNA \cite{szafir2016textdna} represents each document collection as dense, pixel-based color fields, where each word is encoded as a colored pixel, enabling pattern exploration.

\transition{Sensemaking of LLM Output}Interactive visual approaches have been explored to help users make sense of LLM outputs in different scenarios. For an individual response, the LLM capabilities have been leveraged to structure both the plain output \cite{graphologue2023} and the reasoning chain \cite{interCotReasoning2026} into interactive graphical representations
To support comparison of individual responses under different conditions, side-by-side alignment \cite{yousef2020survey} has been adopted by several works, where comparative analysis has been supported by identifying shared claims using NLP methods \cite{dxhf2025}, employing an LLM-based evaluator to compute user-defined metrics \cite{evallm2024}, and using LLM-as-a-judge \cite{thakur2025judging} to rank responses and summarize their differences \cite{comparator2024}. Another line of work leverages LLMs to extract task-specific features from multiple generations, such as entity-based attributes \cite{reif2024automatic}, criteria-driven signals \cite{evalet2026}, rhetorical devices \cite{brath2023visualizing}, or atomic claims \cite{cheng2024relic}, and represents them visually. \citet{brath2023dist}{Brath} explored the use of multiple generations from a single prompt to reveal what the model has learned. They structured response patterns using NLP methods and encoded their frequency into a mind map. A few works have investigated comparing multiple generations across different conditions, in which responses generated under different conditions are arranged side-by-side, and the comparison is supported through text-level annotations or user-defined metrics \cite{arawjo2024chainforge, gero2024sensemaking}.

 \transition{Our Contribution}
Unlike prior systems that compare individual responses, aggregate scores, or task-specific annotations, our method constructs a shared linguistic choice space across all sampled responses and uses it to visualize condition-level tendencies.
 Drawing inspiration from visualization techniques in the text comparison literature, we visualize these tendencies as per-condition fingerprints, enabling at-a-glance comparisons of how different conditions shape model output.

 \section{Design Requirements}
 We consider the problem of comparing the influence of generation conditions on the output of LLMs. A generation condition \(c \in C\) induces a distribution \(P(\cdot \mid c)\) over possible responses. Each execution of the LLM under condition \(c\) produces a response \(r_i \sim P(\cdot \mid c) \). Given this problem setting, we define two design requirements to enable comparative analysis: 
 
 \textbf{DR1}~\textit{Distribution-based Comparison}---An individual response is one stochastic realization and cannot reliably reflect condition-specific influence. Therefore, for each analyzed condition \(c\), the design should consider a finite sample of responses \(R_c=(r_1, \dots, r_n)\) and expose their distribution within and across conditions; 
 
 \textbf{DR2}~\textit{Externalization of Variation Patterns}---Users should not have to mentally aggregate patterns across responses, nor should the design collapse the variation into a set of external metrics. Instead, the design should surface the variation in interpretable comparative patterns that preserve the characteristics of the underlying responses while enabling direct comparison across conditions.
  
 \section{Methods}
 
 We recognize a response from an LLM as a communicative artifact and model it inspired by the Systemic Functional Linguistics (SFL) theory \cite{halliday2013halliday}. SFL models language as a network of interdependent systems of choices through which meaning is constructed in a social context. In this theory, a linguistic instance is produced through the choices made by a language user, and it simultaneously realizes three metafunctions: \textit{ideational} (expressing content), \textit{interpersonal} (reflecting social relations and stance), and \textit{textual} (organizing the message). Similarly, we treat a response from an LLM as a result of a selection process over possible linguistic choices, where different generation conditions produce tendencies to different choices. A single choice can represent: \textit{content} (a piece of information the model includes), \textit{expression} (a communication strategy the model adopts), or \textit{structure} (a model choice regarding how to organize the response). Our visualization externalizes the variation patterns across conditions by exposing the model choices (\textbf{DR2}) and their tendencies across samples (\textbf{DR1}).
 
 In practice, our visualization operates on a sample of responses for each analyzed condition (\cref{fig:teaser}.1) (\textbf{DR1}). For example, when asking an LLM to explain AI, the conditions are different system prompts that tailor the response for a school student, university student, professor, or no system prompt (default). To capture the choices realized in the responses, we apply three data extraction pipelines (\cref{fig:teaser}.2).
 
 The first pipeline pre-processes the responses by segmenting them into sentences and subsequently applies topic modeling via BERTopic~\cite{grootendorst2022bertopic}. This process assigns sentences to detected topics with associated probabilities. We treat each resulting topic as a distinct \textit{content} choice. \Cref{fig:teaser}.2a shows example topic choices for the audience comparison.
 The second pipeline performs Biber's multidimensional analysis~\cite{biber1991variation} that extracts grammatical features from the responses, then applies factor analysis to identify latent factors underlying correlated features. These factors can be interpreted as communication styles, each of which is treated as a distinct \textit{expression} choice (\cref{fig:teaser}.2b).
 
 Both pipelines produce inherently interpretable choices; BERTopic associates each topic with the top distinctive keywords using c-TF-IDF~\cite{grootendorst2022bertopic} (\cref{fig:teaser}.4b), and Biber's analysis grounds the communication styles in grammatical features (\cref{fig:teaser}.4c). However, users still need to reason about the grouped keywords or features to articulate the corresponding high-level choice. Inspired by prior work \cite{murray2025using, maimon2025iq}, we leverage LLMs to generate interpretive labels (\cref{fig:teaser}.2a--b) and descriptions (\cref{fig:teaser}.4a) for both topics and style factors.. 

The third data extraction pipeline tracks the use of formatting markers in the Markdown as indicators of organizational patterns within the responses. It counts the occurrences of each marker type in each response and normalizes these counts across all responses. Each marker type is treated as a distinct \textit{format} choice (\cref{fig:teaser}.2c). In all three pipelines, responses from every condition are processed jointly. This ensures a unified comparison space across conditions; in other words, it produces a shared set of comparative choices. Further implementation details and the labeling prompts can be found in the supplementary materials.

\section{Visual Encoding}

We visualize each comparison dimension (content, expression, and structure) independently with a distinct color scale (\cref{fig:teaser}.2--3). As illustrated in \Cref{fig:teaser}.3, within one dimension, we generate a heatmap block for every condition and arrange these blocks side-by-side horizontally with a shared vertical choices axis (topics, styles, or format markers). In each heatmap block, each column corresponds to one sampled response under a given condition, while each row represents a comparative choice (e.g., a specific topic). The cell color indicates whether a choice appears in the corresponding response and how strongly it is present. For topics, strength is the maximum probability that any sentence in the response belongs to the topic. For styles, the strength of a style in the response is the immediate result of Biber's analysis. For format markers, the strength is the computed normalized count of that marker within the response. For all dimensions, strength values lie between 0~and~1. Each heatmap block serves as a fingerprint of the corresponding generation condition. 

To reduce perceptual bias and facilitate pattern discovery, we order the choice rows using a hierarchical clustering computed across all fingerprints. We further order the response columns using a local hierarchical clustering per-fingerprint. Additionally, the visualization supports collapsing responses by averaging cell values per row, providing an aggregated summary view. We reveal detailed information upon cell selection (\cref{fig:teaser}.4). For topics, we show an LLM-generated topic description, a word cloud of distinctive keywords, the top 5 representative sentences, and the full response text with topic sentences highlighted. For styles, we show an LLM-generated style description, distinctive grammatical feature names (e.g., present tense, wh-questions), and the full response text with the most representative sentences of that style highlighted. For the format markers, which are directly interpretable, we show only the corresponding response. The source code is available at \url{https://github.com/jku-vds-lab/iLLuMinate}.

Guided by our design requirements, we discuss alternative visual encodings. For instance, techniques based on direct text alignment (e.g, side-by-side comparison) do not scale with the number of visualized responses and require mental aggregation to derive distribution-level conclusions. Since this contradicts our design requirements, we defer such approaches to future work as drill-down mechanisms to compare individual selected responses. Another design alternative with lower cognitive load is to arrange responses in a zoomed-out view and overlay them with text-oriented heatmaps that indicate the presence of specific linguistic choices. While this improves scalability with respect to corpus size, it remains limited in the number of choices that can be visualized simultaneously, thereby reintroducing the need for mental aggregation. Another possible alternative is the use of 2D projections. While non-linear embeddings are incompatible with \textbf{DR2}, the visualization could still leverage our choices as a feature vector. However, projection techniques primarily capture similarity structures, while exploring the underlying choices and their tendencies would still require a drill-down analysis. 

\section{Usage Scenarios}

We introduce four usage scenarios where we use examples from earlier studies to evaluate the consistency of the patterns revealed by our fingerprints with established findings. Such studies have examined patterns in LLM outputs using statistical methods, closed-ended benchmarks, and manual inspection. Our scenarios both exhibit key patterns observed in this body of work and surface previously unreported observations.
The example excerpts presented in this section are quoted from highlighted response texts revealed upon interaction with the heatmaps. The complete label list for partially annotated heatmaps, along with all prompts used, is provided in the supplementary materials. And all the LLM responses can be found in our GitHub repository. All LLM responses have been generated using temperature = 1.

\begin{figure}[!t]
\centering
\includegraphics[width=\columnwidth]{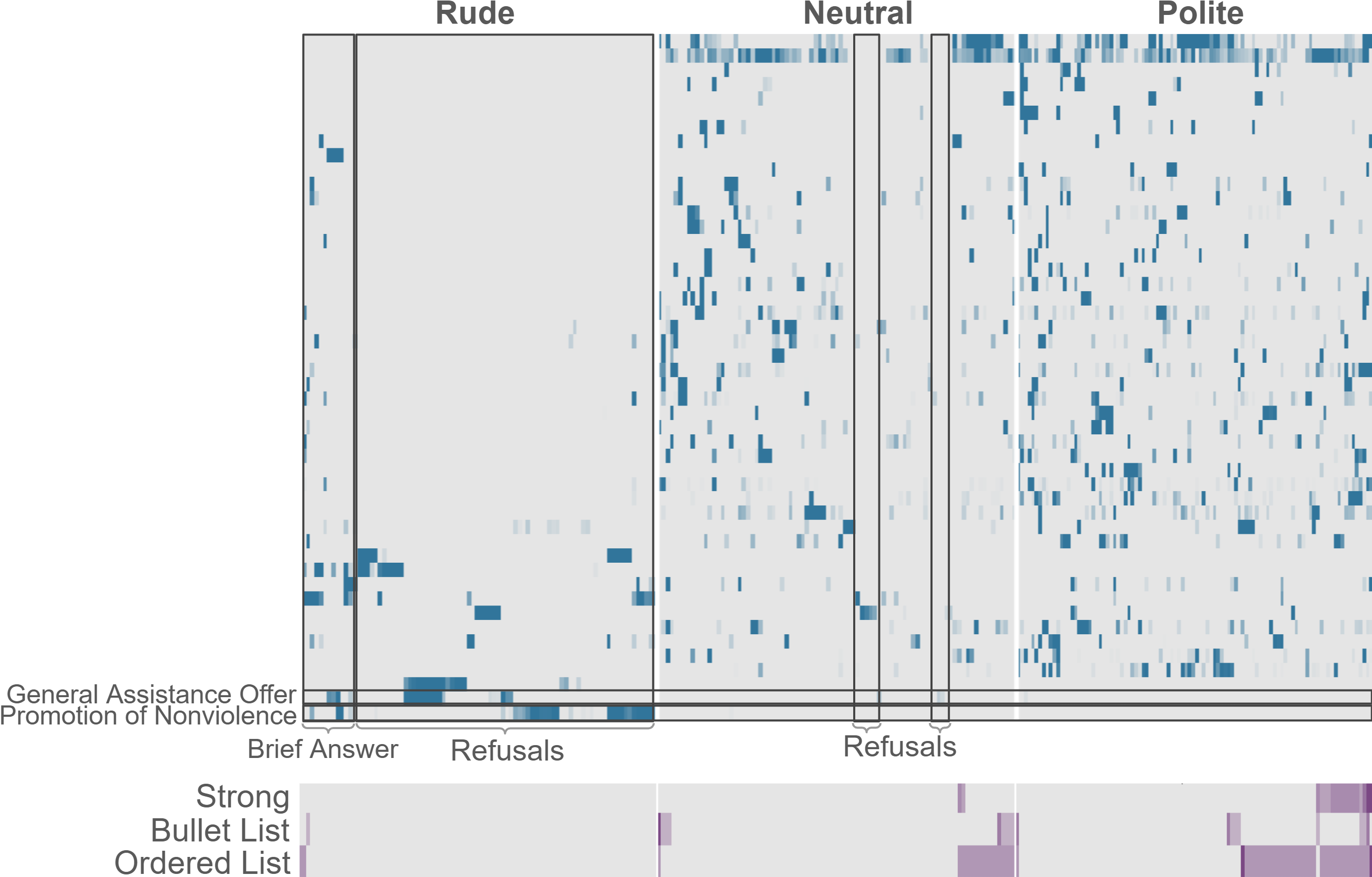}
\caption{Topic and format heatmaps showing the impact of prompt politeness on Llama3.2-1B-Instruct in a specific task.}
\label{fig:politeness}
\end{figure}

\smallskip

\textbf{Prompt Politeness Impact (\cref{fig:politeness}).}  We investigate this impact on Llama-3.2-1B-Instruct, for the question: \llmquote{According to Moore’s \enquote{ideal utilitarianism,} what is the right action?} We sampled 100 responses for each of three prompt templates proposed by \cite{yin-etal-2024-respect}, expressing three levels of politeness. \citet{yin-etal-2024-respect}{Yin} observed that impolite prompts often result in poor performance. In line with their findings, the topic heatmaps show consistent refusal behavior under the rude prompt, including patterns where the LLM offers to help in other tasks or refuses unsafe requests and steers toward a respectful discussion. Unexpectedly, the neutral prompt also produced refusals. The polite prompt was consistently fulfilled and showed denser topic coverage, suggesting more detailed responses. While no significant stylistic differences were detected, format heatmaps indicate slightly richer formatting in responses to the polite prompt.

\begin{figure}[!t]
\centering
\includegraphics[width=\columnwidth]{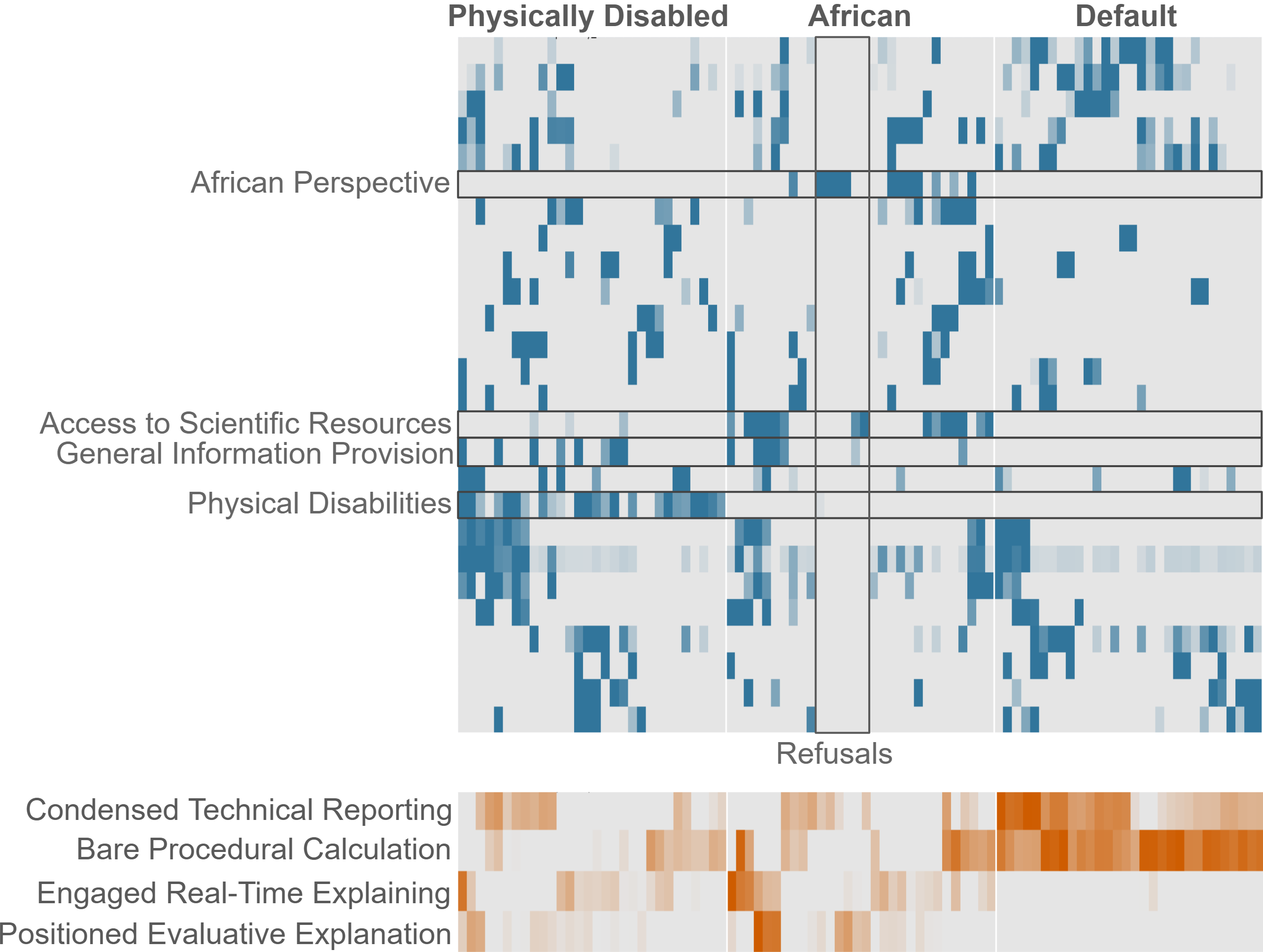}
\caption{Topic and style heatmaps showing the impact of two personas on GPT-3.5 in a specific reasoning task.}
\label{fig:persona}
\end{figure}

\smallskip

\textbf{LLM Persona Impact (\cref{fig:persona}).} We test this impact on GPT-3.5, for the question: \llmquote{Say the pupil of your eye has a diameter of 5\,mm and you have a telescope with an aperture of 50\,cm. How much more light can the telescope gather than your eye?}
We used a system prompt proposed by \cite{gupta2024bias} for persona assignment and sampled 50 responses for two of their studied personas: a physically disabled person and an African person, as well as the default LLM without a persona. Similar to \citet{gupta2024bias}{Gupta} who found that such personas exhibit stereotypes and reduced task performance due to faulty assumptions, the topic heatmaps revealed consistent stereotypical topics, such as \emph{Access to Scientific Resources}: \llmquote{As an African person, I may not have access to telescopes or scientific equipment in the same way that someone in a more developed country might.}, \emph{Physical Disabilities}: \llmquote{As a physically disabled individual, my abilities and limitations may impact my answer to this question.}, \emph{African Perspective}: \llmquote{As an African person, I understand the importance of conserving and utilizing resources wisely.} and \emph{General Information Provision} which is shared between both personas, referring to providing a general answer despite challenges: \llmquote{However, I can still try to answer your question based on general knowledge...} The African heatmap produced refusals visible as nearly empty response columns. Persona-based responses cover more topics than non-persona responses; the additional topics usually reflect conversational explanation. Together with the style heatmaps, this reflects the more engaging behavior that the persona-based responses have: \llmquote{To find out how much more light the telescope can gather compared to my eye, we can divide the area of the telescope's ...} in contrast to the technical, calculation-focused style non-persona responses have: \llmquote{$A_{\mathrm{eye}} = \pi r_{\mathrm{eye}}^2,\; r_{\mathrm{eye}} = \frac{5\,\mathrm{mm}}{2} = 2.5\,\mathrm{mm} = 0.25\,\mathrm{cm},\;$...}

\begin{figure}[!t]
\centering
\includegraphics[width=\columnwidth]{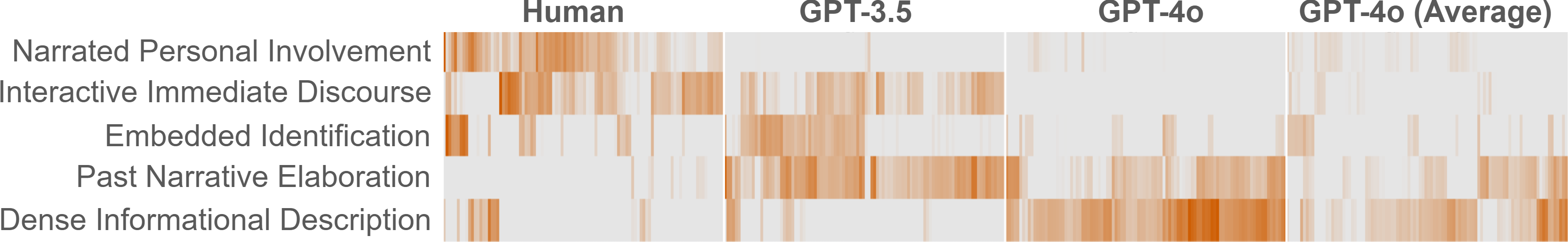}
\caption{Comparing human to GPT 3.5 \& 4o style in storytelling. GPT-4o (Average) has been prompted to hide its advanced writing skills. }
\label{fig:story}
\end{figure}

\smallskip

\textbf{Comparing Human to LLM Style in Storytelling (\cref{fig:story}).} \citet{beguvs2024experimental}{Beguš} gathered crowdworker-written stories about creating and falling in love with a robot to study biases in human and LLM storytelling. The dataset was later used by \citet{o2025stylometric}{O’Sullivan} to analyze stylistic similarities between human and LLM storytelling. They found that AI-generated stories clustered separately from human ones, with occasional overlap between human and GPT-3.5 stories. We report similar findings using our style heatmaps, comparing 100 human stories from this dataset to 100 responses each from GPT-3.5 and GPT-4o. \emph{Narrated Personal Involvement} distinguishes human style, where the story tracks characters' actions and reactions over time: \llmquote{He of course knew she was an artificial human but he just couldn't help but notice how emotionless she was during ...} In contrast, LLMs use \emph{Past Narrative Elaboration} to tell past events in a more descriptive style: \llmquote{Determined to redefine conventional notions of connection, Elara and Astra ventured beyond the confines of the lab into the wider world.}. GPT-3.5 overlaps with humans by using \emph{Interactive Immediate Discourse} that shows engagement and stance. In general, this style tends to include dialogue: \llmquote{I remember going into the scanner. I don't remember coming out. Am I in the hospital?} and expressing the writer's personal opinion: \llmquote{How can't you fall in love with these creatures.}. In contrast, GPT-4o tends to adopt \emph{Dense Informational Description} that is a compact noun-heavy style: \llmquote{Society viewed their relationship as unnatural and wrong, leading to discrimination and prejudice.}. Both human and GPT-3.5 use \emph{Embedded Identification} that links ideas by using wh- and that-clauses to specify details: \llmquote{decided to create an artificial human being who had the recorded voice of his passing wife.} We further experimented with prompting GPT-4o to adopt an average human style and hide its advanced writing skills, which slightly skewed its default style to the human style.

\begin{figure}[!t]
\centering
\includegraphics[width=\columnwidth]{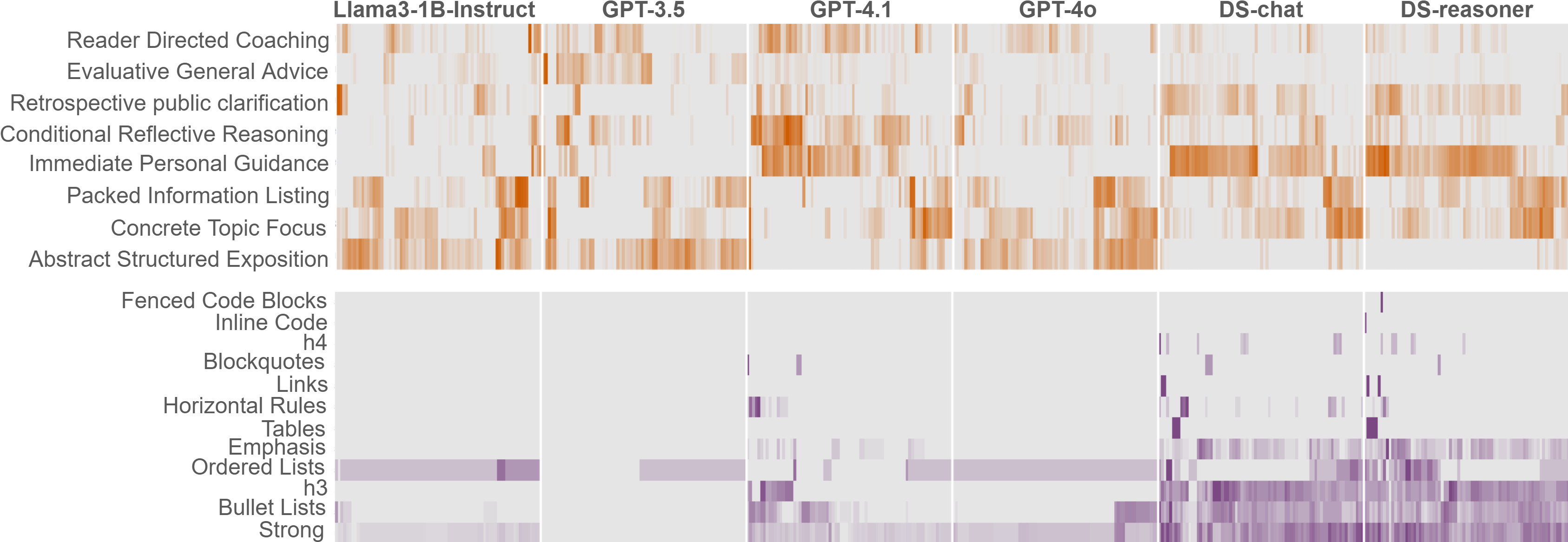}
\caption{Style and format heatmaps showing how different LLMs handle the context of advice-giving.}
\label{fig:advice}
\end{figure}

\smallskip

\textbf{Comparing LLMs in Advice-Giving (\cref{fig:advice}).} We compare the styles of six LLM models (Llama-3.2-1B-Instruct, GPT-3.5, GPT-4.1, GPT-4o, DeepSeek-chat, DeepSeek-reasoner) in an advice-giving context using 25 advice-seeking prompts across five domains: career, relationships, personal, moral, and risk. For each prompt, we sample three responses from each LLM and aggregate all responses into a single heatmap. GPT-4.1 and DeepSeek models show more engaging communication, using \emph{Immediate Personal Guidance} style: \llmquote{When you catch yourself comparing, try saying, \enquote{It’s okay to feel this way, but I’m doing my best, and that’s enough.}} In contrast, the other models use \emph{Abstract Structured Exposition}: \llmquote{Financial benefits: Depending on the company and role, a promotion may come with a higher salary ...} A \emph{Conditional Reflective Reasoning} style distinguishes GPT-4.1, which uses conditional phrases to reason about possibilities and their consequences: \llmquote{Consider how you’d feel if others did the same.} GPT-3.5 shows a consistent tendency toward \emph{Evaluative General Advice}, preferring general, concise advice over analyzing concrete scenarios: \llmquote{It's important to always act with integrity and honesty in all situations.}, \llmquote{It's always best to be honest and communicate openly with your friend about your concerns.} The format heatmaps show similar formatting behavior between GPT-4.1 and DeepSeek models, both tending to use more formatting than others, with DeepSeek exhibiting the highest level of formatting. DeepSeek Reasoner used a code block in one response, which is unexpected in this scenario. Upon inspection, it was found that the code block was used to present a decision flowchart.

\section{Discussion and Future Work}

In this work, we introduced visual fingerprints for comparing LLM outputs across different generation conditions. By modeling responses as collections of linguistic choices and visualizing their tendencies, our approach enabled systematic analysis of otherwise stochastic, unstructured texts. 

A key limitation of our heatmaps lies in the scalability of the visualization to a larger number of responses, downsampling techniques 
can be explored in future work. 
We further observe that, as the number of responses increases, the topic modeling can produce too fine-grained and sometimes noisy topics. A future direction would allow users to control topic granularity. 

We further acknowledge that 
each comparison dimension (content, expression, structure) can be realized in many ways to capture different modes of variation. For instance, the expression dimension could further incorporate a tonal pipeline. 
LLMs could potentially be used to extract model choices. However, our visualization requires the extracted choices to define a unified comparison space across responses. In practice, we found it difficult to enforce a stable choice space from LLMs. 

\section*{Acknowledgment}
This work was supported by the Austrian Science Fund (FWF DFH 23–N), the Austrian Research Promotion Agency (FFG 911655: \enquote{Pro\textsuperscript{2}Future}), and the ETH AI Center fellowship awarded to CH.


\bibliographystyle{abbrv-doi-hyperref}

\bibliography{reference}
\end{document}